\documentclass[conference]{IEEEtran}
\IEEEoverridecommandlockouts

\usepackage{cite}
\usepackage{amsmath,amssymb,amsfonts}
\usepackage{algorithmic}
\usepackage{graphicx}
\usepackage{textcomp}
\usepackage{xcolor}
\usepackage{tikz}
\usepackage{algorithmic,algorithm,stmaryrd}
\usepackage{amsthm}

\def\BibTeX{{\rm B\kern-.05em{\sc i\kern-.025em b}\kern-.08em
    T\kern-.1667em\lower.7ex\hbox{E}\kern-.125emX}}

\theoremstyle{definition}

\newcommand{\R}{\mathbb{R}}                      
\DeclareMathOperator*{\argmax}{arg\,max}

\begin{document}

\title{Rare geometries: revealing rare categories via dimension-driven statistics}

\author{\IEEEauthorblockN{Henry Kvinge*\thanks{*Corresponding author}}
\IEEEauthorblockA{\textit{Department of Mathematics} \\
\textit{Colorado State University}\\
Fort Collins, CO, USA\\
henry.kvinge@colostate.edu}
\and
\IEEEauthorblockN{Elin Farnell$^\dagger$\thanks{$^\dagger$Elin Farnell's contributions to this work were completed while she was a research scientist in the Department of Mathematics at Colorado State University.}}
\IEEEauthorblockA{\textit{Amazon} \\
Seattle, WA, USA\\
efarnell@amazon.com}
\and
\IEEEauthorblockN{Jingya Li}
\IEEEauthorblockA{\textit{Department of Mathematics} \\
\textit{Colorado State University}\\
Fort Collins, CO, USA\\
jingyaaa@rams.colostate.edu}
\and
\IEEEauthorblockN{Yujia Chen}
\IEEEauthorblockA{\textit{Department of Mathematics} \\
\textit{Colorado State University}\\
Fort Collins, CO, USA\\
ujchen@rams.colostate.edu}
}

\maketitle

\begin{abstract}
In many situations, classes of data points of primary interest also happen to be those that are least numerous.
A well-known example is detection of fraudulent transactions among the collection of all financial transactions, the vast majority of which are legitimate. These types of problems fall under the label of `rare-category detection.' There are two challenging aspects of these problems. The first is a general lack of labeled examples of the rare class and the second is the potential non-separability of the rare class from the majority (in terms of available features). Statistics related to the geometry of the rare class (such as its intrinsic dimension) can be significantly different from those for the majority class, reflecting the different dynamics driving variation in the different classes. In this paper we present a new supervised learning algorithm that uses a dimension-driven statistic, called the kappa-profile, to classify whether unlabeled points belong to a rare class. Our algorithm requires very few labeled examples and is invariant with respect to translation so that it performs equivalently on both separable and non-separable classes.
\end{abstract}


\author{\IEEEauthorblockN{Henry Kvinge, Elin Farnell, Jingya Li, Yujia Chen}
\IEEEauthorblockA{Department of Mathematics\\
Colorado State University\\
Fort Collins, CO 80523-1874}
}

\section{Introduction}
\label{sec:introduction}

Rare-category detection is a common problem in real-world settings where it is often the case that classes that are most important to identify are least well-represented in available datasets. In such cases, we may have only a handful of labeled examples of the rare class even though we have many labeled examples of the majority class. Typical examples include identification of financial fraud, malicious insiders in an organization, fraudulent transactions, rare diseases, and unusual objects that are not explained by current models in astronomy; see, e.g. ~\cite{dokas2002data,pelleg2005active,bay2006large,he2008nearest,he2010rare}. 

Past approaches have included use of a mixture model \cite{pelleg2005active}. Other works use the density of unlabeled points around a point known to belong to the rare class in order to to detect a cluster from the rare class. For example, in \cite{he2008nearest} a local-density-differential-sampling strategy was used. In \cite{he2008graph} on the other hand, a graph-based approach using similarity matrices was used to capture changes in density.

One key distinction between the rare-category detection problem and the problem of, for example finding outliers of a dataset, is that a rare class is generally not assumed to actually be separable from the majority class (at least in the given features). This makes classification challenging, and we are forced to rely heavily on the small number of labeled points which we know belong to the rare class. When points from such rare classes are not separable (at least with their given features) from majority classes, they often have other characteristics that can help us to identify whether an unlabeled point is likely to belong to the rare class. In this paper we are interested in cases where rare and majority classes have different geometry or ``shape.'' Such a situation is plausible when the distribution and variance of the majority and rare classes are driven by different processes. In the most extreme (but not unusual) case, two classes will have different geometries if they have different intrinsic dimensions. 

We use a statistic called the $\kappa$-profile \cite{kvinge2018monitoring} which can be calculated for a set of points $D \subset \R^n$. Very roughly, the $\kappa$-profile measures how well $D$ can be projected into a range of subspaces of varying dimension in such a way as to best satisfy an optimization problem \eqref{eqn-opt-problem}. Among other things, it is an effective tool for estimating the intrinsic dimension of a dataset. In our algorithm, which we call the $\kappa$-detection algorithm, we use a comparison of the $\kappa$-profile of a cluster of rare class points against the $\kappa$-profile of the same cluster with the inclusion of an unlabeled point, as a metric by which to determine whether that unlabeled point belongs to the rare class. The underlying assumption is that even if an unlabeled point is very close to a cluster of rare class points, if it does not agree with the geometry of this cluster then it is probably not a point from the rare class. Because the $\kappa$-detection algorithm draws heavily from characteristics related to the intrinsic dimension of a dataset we expect it to actually perform better on data sampled from a higher ambient dimension because in higher dimensions there is likely more dimension-related information to extract.

This paper is structured as follows. In Section \ref{sec:background}
 we summarize background information on secant-based dimensionality-reduction algorithms, the concept of a $\kappa$-profile, and the context and set of assumptions we make in the rare-category detection problem. In Section \ref{sec:kappa-detection} we present the $\kappa$-detection algorithm and a method for determining a key threshold parameter in this algorithm. In Section \ref{sec:examples} we apply the $\kappa$-detection algorithm to both real and synthetic examples. Finally in Section \ref{sec:conclusion} we describe some future directions.

\section{Background}
\label{sec:background}

\subsection{Secant-based dimensionality reduction}

In most applications, a reasonable dimensionality-reduction algorithm should preserve the distance between two data points in their ambient space. Such a goal can be equivalently stated as the requirement that the secant set $S$ of a dataset $D \subset \mathbb{R}^n$ be preserved during dimensionality reduction. In this paper we will choose to work with the normalized secant set $S$, 
\begin{equation*}
    S := \Big\{ \frac{x - y}{||x-y||_{\ell_2}} \;\Big|\; x,y \in D, x \neq y \Big\}.
\end{equation*}
The purpose of normalization is to give equal footing to both large- and small-scale structure. When working with real data, it is often useful to discard the very smallest secants as these are most affected by noise. 

The dimensionality-reduction algorithms which underlie this paper all attempt to solve the secant-based optimization problem: 
\begin{equation} \label{eqn-opt-problem}
    \argmax_{P \in \text{Proj}(n,k)} \min_{s \in S} ||P^Ts||_{\ell_2}.
\end{equation}
Here $\text{Proj}(n,k)$ is the collection of all $n \times k$ matrices (with $k \leq n$) whose columns are orthonormal. Note that this set is equivalent to the set of orthogonal $k$-projections from $\R^n$ to $\R^n$. Roughly, \eqref{eqn-opt-problem} attempts to find the projection onto a $k$-dimensional subspace such that the length of the secant $s$ which is least well-preserved is maximized. This is in contrast to principal component analysis (PCA) for example, which solves a different optimization problem. As a result, a solution to \eqref{eqn-opt-problem} frequently differs from the corresponding PCA solution. Problem \eqref{eqn-opt-problem} is closely tied to the intrinsic dimension of $D$ via the constructive proof of the Whitney Embedding Theorem from differential topology \cite[Theorem 6.15]{lee2013smooth}. This fact will be a key aspect of the algorithm proposed in this paper.

There are fast, lightweight, iterative algorithms that converge to local optima for \eqref{eqn-opt-problem} \cite{kvinge2018gpu}, \cite{kvinge2018too}. For the experiments in this paper we used the SAP algorithm from  \cite{kvinge2018gpu}. The SAP algorithm is well-adapted to working with rare categories because it scales well to high dimensional data though less well to large numbers of points. In the case of rare categories the latter is not an issue by assumption.

\subsection{The $\kappa$-profile} \label{sec:kappa-profile}

In this section we review the concept of $\kappa$-values and $\kappa$-profiles. The $\kappa$-profile statistic will be the primary tool that we will use to determine whether an unlabeled point conforms to the known geometry or ``shape'' of a rare class. 

The notion of a $\kappa$-value was first defined in \cite{BK00}. Such values arise from solutions to \eqref{eqn-opt-problem}. Specifically, let $P^*$ be the projection that satisfies \eqref{eqn-opt-problem} for some projection dimension $k\leq n$; then the $\kappa$-value $\kappa_k$ is defined as
\begin{equation*}
    \kappa_k := \min_{s \in S} ||P^*s||_{\ell_2}.
\end{equation*}
Note that because we assume that the elements of our secant set $S$ have been normalized, it is always the case that $\kappa \in [0,1]$. Suppose that $\mathbf{k} = (k_1,k_2, \dots, k_m)$ is an increasing sequence of integers such that for each $i,$ $1 \leq k_i \leq n$ (we will generally assume that the $k_i$'s are consecutive but they need not be). Then the $\kappa$-profile associated with $\mathbf{k}$ is the tuple of $\kappa$ values
\begin{equation*}
    \pmb{\kappa}_{\mathbf{k}} := (\kappa_{k_1}, \kappa_{k_2}, \dots, \kappa_{k_m}).
\end{equation*}
In analogy to the singular values produced when applying PCA, the $\kappa$-profile tells us something about how well our dataset can be projected into lower-dimensional spaces. The information provided by the $\kappa$-profile however is more sensitive to the intrinsic dimension of the dataset (see Section II.C in \cite{kvinge2018monitoring}).

\begin{figure}
\begin{center}
\includegraphics[height=6.2cm]{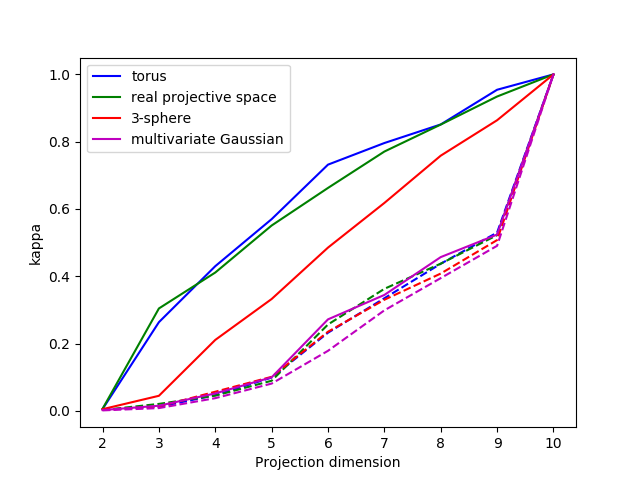}
\end{center}
\caption{\label{fig:SRreconstructions}} The $\kappa$-profiles for 100 points drawn from: a torus (solid blue), the real projective plane $\mathbb{RP}^2$ (solid green), the 3-sphere (solid red). All these were smoothly mapped into $\R^{10}$. For comparison we also plot the $\kappa$-profile for a collection of points drawn from a multivariate Gaussian distribution in $\R^{10}$ (solid magenta). We add a random point from $\mathbb{R}^{10}$ to each collection of points and plot the new $\kappa$-profile with a dashed line. As can be seen, adding a single point away from the embedded manifold results in a $\kappa$-profile that is significantly different from the original, except in the case of the multivariate Gaussian distribution in $\R^{10}.$
\end{figure}

In Figure \ref{fig:SRreconstructions} we plot the $\kappa$-profiles for points drawn from several different manifolds (shown as solid curves in the figure), where all manifolds are smoothly mapped into $\R^{10}.$ Specifically, in Figure \ref{fig:SRreconstructions}, we show
\begin{itemize}
\item the $\kappa$-profile for a set $D_{torus}$ of points drawn randomly from a $2$-dimensional torus $T$ mapped smoothly into $\R^{10}$.
\item The $\kappa$-profile for a set $D_{\mathbb{RP}^2}$ of points drawn randomly from the real projective plane $\mathbb{RP}^2$ and mapped smoothly into $\R^{10}$.
\item The $\kappa$-profile for a set $D_{S^3}$ of points drawn randomly from the 3-sphere and mapped smoothly into $\R^{10}$.
\item The $\kappa$-profile for a set $D_{Gaus}$ of random Gaussian noise in $\R^{10}$.
\end{itemize}
As can be seen, the relationship between the  $\kappa$-profiles in this figure reflect the intrinsic dimension of the manifolds from which each set of points was drawn. The torus and $\mathbb{RP}^2$ are both 2-dimensional manifolds and this is reflected by the fact that the $\kappa$-values for the associated sets of points grow the fastest. On the other hand, the $\kappa$-values for the 3-sphere (which is a 3-dimensional manifold) grow more slowly. Finally, the set of points drawn from the multivariate Gaussian distribution in $\R^{10}$ grows the slowest reflecting the fact that the intrinsic dimension of this dataset really is $10$. In general, the $\kappa$-profile for a set of points with lower intrinsic dimension should sit above the $\kappa$-profile for a set of points with higher intrinsic dimension. 

Because the $\kappa$-profile is sensitive to changes in dimension, if we sample a point $x$ randomly from $\mathbb{R}^{10}$ and include this in any of $D_{torus}$, $D_{\mathbb{RP}^2}$, or $D_{S^3}$, we should expect the $\kappa$-profiles of $D_{torus}\cup \{x\}$, $D_{\mathbb{RP}^2}\cup \{x\}$, and $D_{S^3}\cup \{x\}$ to be noticeably different from the original $\kappa$-profile for $D_{torus}$, $D_{\mathbb{RP}^2}$, or $D_{S^3}$. To see this, compare the solid and dashed lines of each color in Figure \ref{fig:SRreconstructions}. The dashed lines are exactly the $\kappa$-profiles of $D_{torus}\cup \{x\}$, $D_{\mathbb{RP}^2}\cup \{x\}$, $D_{S^3}\cup \{x\}$, $D_{Gaus}\cup \{x\}$. As can be seen, just adding a single point which lies off  the original manifold gives a $\kappa$-profile that better matches that of Gaussian noise rather than the original $\kappa$-profile for the manifold.

\subsection{The rare-category detection problem}

In our version of the rare-category detection problem, we assume that we are given a dataset
\begin{equation*}
    X = X_{maj} \cup X_{rare} \cup Y
\end{equation*}
where $X_{maj}$ consists of labeled points known to belong to a majority class, $X_{rare}$ consists of labeled points known to belong to a rare class, and $Y$ consists of unlabeled points. The goal of our algorithm will be to classify whether each point in $Y$ belongs to the rare class or the majority class. 

The algorithm that we describe in Section \ref{sec:kappa-detection} is designed to cope with two of the major challenges of this problem.
\begin{enumerate}
\item The number of training points belonging to the rare class $X_{rare}$, which we train on, is potentially very small (for example, less than $12$).
\item The classes may not be separable in the given features.
\end{enumerate}

\section{The $\kappa$-detection algorithm} 
\label{sec:kappa-detection}

In this section we describe the {\emph{$\kappa$-detection algorithm}}, which utilizes the $\kappa$-profile described in Section \ref{sec:kappa-profile} in order to classify unlabeled points as either belonging to the majority class or the rare class.

The basic idea behind this algorithm is that one way to gauge whether an unlabeled point $y$ belongs to a rare class is whether its inclusion in the rare class substantially changes the class's geometry. Our proxy statistic to detect whether $y$ ``changes the geometry'' of the rare class is the $\kappa$-profile. Specifically, we assume we have been handed a set $X_{rare}$ of labeled rare class points and a set $Y$ that we want to classify as either belonging to $X_{rare}$ or not. We start by calculating the $\kappa$-profile $\pmb{\kappa}_{rare}$ of $X_{rare}$ without including any unlabeled points. We next iterate through each point $y \in Y$ and calculate the $\kappa$-profile $\pmb{\kappa}_y$ of $X_{rare}\cup\{y\}$. We calculate the change $d_y$ in the $\kappa$-profile using the $\ell_2$-norm
\begin{equation} \label{eqn-kappa-shift}
    d_y := ||\pmb{\kappa}_{rare} - \pmb{\kappa}_y||_{\ell_2}.
\end{equation}
If $d_y$ is below a user-specified threshold $thresh$ we label $y$ as a point in $X_{rare}$; otherwise we label it as a majority point. The algorithm as a whole is outlined in Algorithm \ref{algo-kappa-detection}.

There are several parameters which need to be tuned for Algorithm \ref{algo-kappa-detection} to perform well. The first is the threshold, $thresh$. In Section \ref{sec:algorithmic_thresholds} we propose a data-driven algorithm for determining $thresh.$ Of course, the appropriate choice of threshold may differ depending on the application. In some applications it is more important to avoid false positives while in other situations false negatives are worse. In the former case we should pick a smaller threshold and in the latter we should pick a larger threshold.

The optimization problem \eqref{eqn-opt-problem} is non-convex and therefore one is unlikely to actually find the global solution $P^*$. Which maximum is found is based on the initial projection that is used as well as the step-size in the SAP algorithm (for a discussion of these parameters, see \cite{kvinge2018gpu}). Because the SAP algorithm is relatively fast, for more accurate approximations of the $\kappa$-profiles $\pmb{\kappa}_{rare}$ of $X_{rare}$ and $\pmb{\kappa}_{y}$ of $X_{rare}\cup\{y\}$, one can choose to compute each of these $t$ times, where $t$ is a parameter chosen by the user; then we take the pointwise average of the $t$ $\kappa$-profiles and call these $\pmb{\kappa}_{rare}$ or $\pmb{\kappa}_y$ respectively. In the experiments in this paper we generally took $1 \leq t \leq 10$.

Finally, this paper rests on the basic assumption that a class $C \subset \R^n$ of points in a dataset approximately sits on a $k$-dimensional manifold embedded in $\R^n$ with $k < n$. This identification is never exact because of noise in the dataset. In terms of secants, this noise will have a much more significant effect on short secants. For this reason, when applying Algorithm \ref{algo-kappa-detection} to real data we advocate discarding the shortest secants when calculating $\kappa$-profiles. 

\begin{algorithm} 
\caption{\label{algo-kappa-detection} The $\kappa$-detection algorithm}
\begin{algorithmic}[1]
\STATE \textbf{inputs} A labeled collection $X_{rare}$, the unlabeled set of points $Y$, the number of trials $t$, the range of dimensions $\mathbf{k}$ for which we will calculate the $\kappa$-profile.
\STATE Calculate the $\kappa$-profile $\pmb{\kappa}_{rare}$ for $X_{rare}$ for the range of dimensions $\mathbf{k}$.
\FOR{$y \in  Y$}
    \FOR{$j \leq t$}
        \STATE Calculate the $\kappa$-profile $\pmb{\kappa}_{y,j}$ of $X_{rare} \cup \{y\}$ from a randomly initialized projection. Possibly discard secants shorter than some user specified length during this calculation.
    \ENDFOR
    \STATE $\pmb{\kappa}_y \mapsfrom$ element-wise average of $\pmb{\kappa}_{y,1}, \dots, \pmb{\kappa}_{y,t}$.
    \STATE Set $d_{y} := ||\pmb{\kappa}_{rare} - \pmb{\kappa}_{y}||_{\ell_2}$.
    \IF{$d_y < thresh$}
        \STATE Classify $y$ as a member of the rare class.
    \ELSE
        \STATE Classify $y$ as a member of the majority class.
    \ENDIF
\ENDFOR
\end{algorithmic}
\end{algorithm}

\subsection{Algorithmic determination of thresholds}
\label{sec:algorithmic_thresholds}

The $\kappa$-detection algorithm requires a threshold value which determines the extent to which a point is allowed to alter the $\kappa$-profile of the rare class before we say that this point is not an element of the rare class. We have left this as a parameter to be tuned by the user because the choice between a higher or lower threshold should be based on the application.

We do, however, present an algorithm, Algorithm \ref{algo-threshold}, that generates a rough threshold (which can be further refined to the specific dataset) based on what is already known about the rare class. The idea of the algorithm is that one should try to understand how much variation in $\pmb{\kappa}_{rare}$ is introduced by each point which we already know belongs to $X_{rare}$. If we find that on average, for each $x \in X_{rare}$ the $\kappa$-profile for $X_{rare}\setminus \{x\}$ is significantly different than the $\kappa$-profile to $X_{rare}$, then we should not be surprised that for $y \in Y$ belonging to the rare class, the $\kappa$-profile of $X_{rare}\cup\{y\}$ might differ significantly from the $\kappa$-profile for $X_{rare}$.

Algorithm \ref{algo-threshold} begins by calculating the $\kappa$-profile for the labeled points from the rare class, $\pmb{\kappa}_{rare}$. Next we iterate through all $x \in X_{rare}$ and for each $x$ we calculate the $\kappa$-profile $\pmb{\kappa}_x$ of $X_{rare}\setminus\{x\}$. We set
\begin{equation*}
    d_x = ||\pmb{\kappa}_{rare} - \pmb{\kappa}_{x}||_{\ell_2}.
\end{equation*}
Finally we take the pointwise average over all $d_x$ for $x \in X_{rare}$ and call this vector $d_{avg}$. We have found that in practice, a good starting threshold is $thresh = rd_{avg}$ where $1.1 \leq r \leq 1.5$. By assumption $X_{rare}$ contains few points, so Algorithm \ref{algo-threshold} is generally fast.

\begin{algorithm} 
\caption{\label{algo-threshold} Algorithmic threshold determination}
\begin{algorithmic}[1]
\STATE \textbf{inputs} A labeled set of rare points $X_{rare} \subset \R^n$, the number of trials $t$ to perform for each $\kappa$-calculation, $1.1 \leq r \leq 1.5$.
\STATE Calculate the $\kappa$-profile $\pmb{\kappa}_{rare}$ of $X_{rare}$.
\FOR{$x \in  X_{rare}$}
    \FOR{$j \leq t$}
        \STATE Calculate the $\kappa$-profile $\pmb{\kappa}_{x,j}$ of $X_{rare}\setminus \{x\}$ from randomly initialized projections.
    \ENDFOR
    \STATE $\pmb{\kappa}_x \mapsfrom$ the pointwise average of $\pmb{\kappa}_{x,1}, \dots, \pmb{\kappa}_{x,t}$.
    \STATE $d_x \mapsfrom ||\pmb{\kappa}_{min} - \pmb{\kappa}_{x}||_{\ell_2}$.
\ENDFOR
\STATE $thresh \mapsfrom r \cdot \text{mean}_{x \in X_{rare}} \{d_x \}$
\end{algorithmic}
\end{algorithm}

\subsection{Limitations}

The $\kappa$-detection algorithm is not without limitations. The most obvious of these is that our algorithm demands at least enough labeled rare class points to estimate the geometry of the rare class. In particular, the number of rare class points must be at least equal to (and preferably more than) $k$, where $k$ is the dimension of the manifold on which $X_{rare}$ approximately sits. The value $k$ is generally not known, however in practice we have found that $k \leq 10$ is a safe assumption for all but the largest and most varied classes. In general one can obtain a reasonably good estimation of the $\kappa$-profile of a dataset even from small subsamples. We conjecture that this is related to the fact that the number of secants grows as $O(p^2)$ as a function of the number of points $p$ in our sample.

The second potential limitation of Algorithm \ref{algo-kappa-detection} is that it will not perform well when the underlying geometry is the same for different classes. Such a phenomenon can arise in cases where class distinctions are artificial. Imagine for example we are trying to label the integer age of adults in a dataset via their physical measurements. The age of 34 might be a rare class, but the underlying dynamics that relate the physical characteristics of a person to their age are probably not very different between individuals who are 33 and those who are 34. On the other hand, the distinction between protein-binding locations on E. coli cells is not artificial and therefore it would not be surprising if the features for different binding locations encode different geometries. See Section \ref{subsec:real_datasets} for performance of the $\kappa$-detection algorithm on a dataset fitting this description.

Because the $\kappa$-detection algorithm utilizes features of data (its geometry for example) that to our knowledge are not used in other rare-category detection algorithms, we believe that it will function particularly well as part of an ensemble of methods. We further expect that by including the strengths of other approaches, the limitations described above will be minimized.

\section{Real and synthetic examples}
\label{sec:examples}

In this section we describe the performance of the $\kappa$-detection algorithm on both synthetic and real data.

\subsection{A synthetic example}

We begin by applying the $\kappa$-detection algorithm to a simple synthetic dataset. This dataset $X$ is the union of two sets of points: points corresponding to the majority class $X_{maj} \subset \R^{6}$ and points corresponding to the rare class $X_{rare} \subset \R^{6}$. The set $X_{maj}$ is itself the union of 6 sets $X_1, X_2, \dots, X_{6}$ where $X_i$ is a set of points drawn from a multivariate normal distribution centered at the origin with covariance matrix equal to a diagonal matrix with entries $0.2$ everywhere on the diagonal except for a value of $1$ in the entry at $(i,i)$. 

\begin{figure}
\begin{center}
\includegraphics[height=6.2cm]{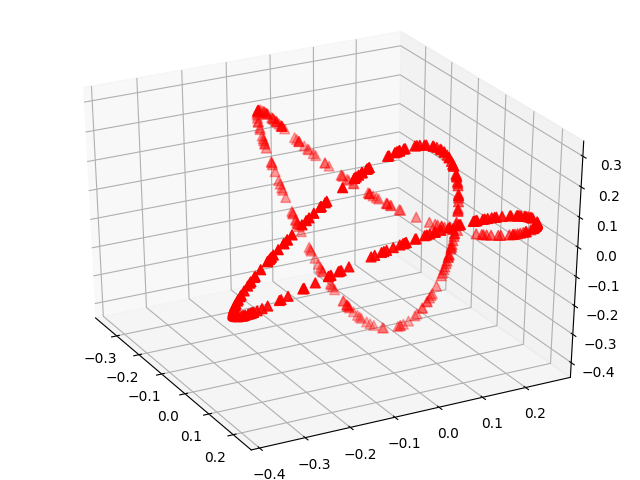}
\end{center}
\caption{\label{fig:trig_curve} A projection of points sampled from the trigonometric moment curve $f(t)$ in \eqref{eqn-trig} into $\mathbb{R}^3$ using PCA. It is clear that this curve is indeed intrinsically $1$-dimensional.} 
\end{figure}

\begin{figure}
\begin{center}
\includegraphics[height=6.2cm]{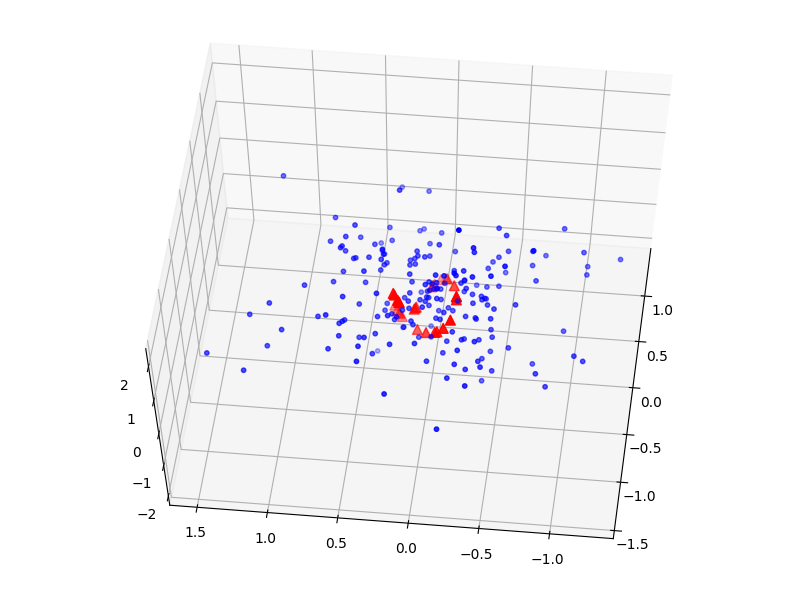}
\end{center}
\caption{\label{fig:synthetic_visualization} A projection of the synthetic dataset $X$ into $3$-dimensions (from $6$-dimensions). The majority class is presented as small blue circles while the points from the rare class are larger, red triangles. Note that while the majority class appears to be more or less a point cloud, the rare class is a 1-dimensional non-linear curve.} 
\end{figure}

The rare class $X_{rare}$ is formed from random points drawn from the $6$-dimensional trigonometric moment curve:
\begin{equation} \label{eqn-trig}
    f(t) := \Big(\cos(t), \sin(t), \dots, \cos(3t), \sin(3t)\Big).
\end{equation}
A projection of points sampled from $f(t)$ is shown in Figure \ref{fig:trig_curve}. As can be seen, the image of $f(t)$ is indeed intrinsically $1$-dimensional. A projection of the whole dataset $X$ into 3-dimensions is shown in Figure \ref{fig:synthetic_visualization}. The blue points are from the majority class, the red points are from the rare class. In this case, for ease of viewing we have not differentiated between points which are labeled and those which are not. 

\begin{figure}
\begin{center}
\includegraphics[height=5.8cm]{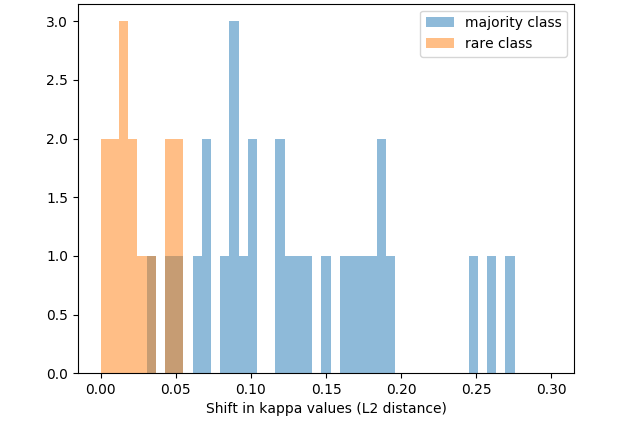}
\end{center}
\caption{\label{fig:histogram_synthetic} A histogram showing the range of values taken by $d_y$ for $y$ either in the rare class itself (orange) or the majority class (blue). As can be seen, $d_y$ is a fairly successful way to identify which class that $y$ belongs to.}  
\end{figure}

In this synthetic example, the intrinsic dimension of the rare class is $1$ (it is a curve) while the intrinsic dimension of the majority class is $6$. Given this significant difference in dimension, we would expect that a random point from the majority class would (even if it is very close to points from the rare class spatially) with high probability be off of the curve $f(t)$ and therefore on average result in a large change in the $\kappa$-profile. This is what we see in Figure \ref{fig:histogram_synthetic}. Here we have plotted a histogram for a range of $d_y$ values (see \eqref{eqn-kappa-shift}) when $y$ is actually a point from the rare class (orange) or when $y$ is a point from the majority class (blue). Clearly if one were to set a threshold of $thresh \approx .05$ one would be able to separate points that come from the rare class and points that come from the majority class reasonably well. 

\begin{figure}
\begin{center}
\includegraphics[height=5.8cm]{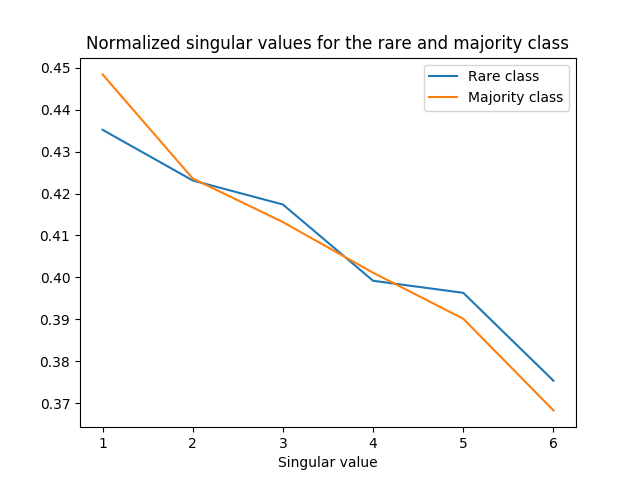}
\end{center}
\caption{\label{fig:singular_synthetic} The normalized singular values for the synthetic rare and majority classes of points. Note that despite the fact that $X_{rare}$ is drawn from a $1$-dimensional manifold and $X_{maj}$ is drawn from a $6$-dimensional manifold, the plot of the singular values of these two datasets look relatively similar.}  
\end{figure}

\begin{figure}
\begin{center}
\includegraphics[height=5.8cm]{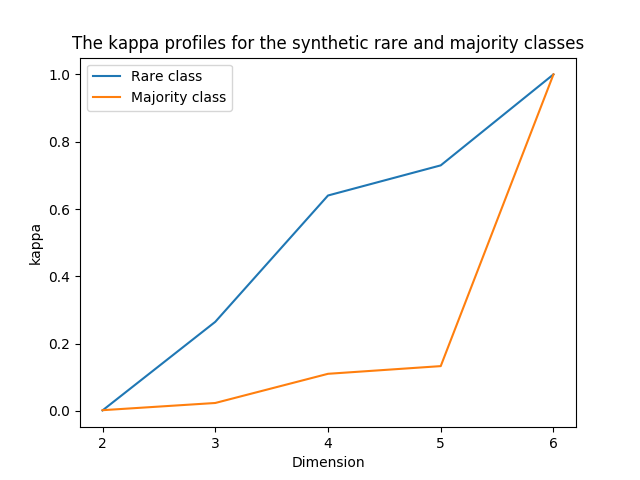}
\end{center}
\caption{\label{fig:kappa_synthetic} The $\kappa$-profiles for the synthetic rare and majority classes of points. Unlike the plot of the singular values in Figure \ref{fig:singular_synthetic}, the $\kappa$-profiles for these two classes actually reflect their intrinsic dimensions.}  
\end{figure}

This synthetic example is also useful for illustrating why solving the optimization problem \eqref{eqn-opt-problem} is essential to the performance of the algorithm. Superficially, it might seem that the $\kappa$-profile could be replaced by singular values. After all, both of these statistics measure how well data can be projected into different dimensions. We plot the singular values for $X_{rare}$ and $X_{maj}$ in Figure \ref{fig:singular_synthetic}. Observe that these curves look very similar despite the fact that $X_{rare}$ is drawn from a $1$-dimensional manifold and $X_{maj}$ is drawn from a $6$-dimensional manifold. On the other hand the $\kappa$-profiles in Figure \ref{fig:kappa_synthetic} look quite distinct and reflect the differences in dimension between $X_{rare}$ and $X_{maj}$. It is easy to see why replacing the $\kappa$-profile in Algorithm \ref{algo-kappa-detection} with singular values would severely limit the algorithm's ability to detect geometry. Only the $\kappa$-profile can detect the dimension despite the non-linearities of the datasets.

\subsection{Real datasets}
\label{subsec:real_datasets}

In this section we apply of the $\kappa$-detection algorithm to four real-world datasets. The results are summarized in Table \ref{fig:performance-table}. Each dataset consists of a number of different imbalanced classes. In each setting, we did the following.
\begin{itemize}
\item We chose a class with relatively few points and called this the `rare class.' We designated the union of all of the rest of the classes as the `majority class.'
\item For all datasets other than the shuttle dataset where we used $thresh = .05$, we used Algorithm \ref{algo-threshold} to determine a threshold. When applying Algorithm \ref{algo-threshold}, we set $thresh=rd_{avg},$ where $1.1 \leq r \leq 1.5$.
\item We ran the $\kappa$-detection algorithm $10$ times for each dataset with a new random partition of the rare and majority classes into labeled and unlabeled points. In Table \ref{fig:performance-table} we record the average percentage of unlabeled rare class points that the algorithm correctly identified, as well as the average percentage of unlabeled majority class points that the algorithm misidentified as rare.
\end{itemize}
As can be seen, very few labeled rare class points (between $8$ and $10$ depending on the dataset) were required to achieve reasonable classification results. Note that in each case, one could improve the values in the second column of Table \ref{fig:performance-table} by decreasing the threshold parameter at the expense of increasing the number of majority points misclassified as rare in the third column. Our implementation was intended to be balanced with respect to this trade-off. Finally, the reader should keep in mind that because the classes are imbalanced in all of these datasets, the percentages in the second and third column of Table \ref{fig:performance-table} can correspond to very different absolute numbers of points. 

\begin{figure*}
\begin{center}
\begin{tabular}{ |c|c|c|c|} 
 \hline
Dataset & \% rare class identified & \% majority class misidentified as rare & \# training points\\ 
\hline
E. coli & 70 & 19 & 9  \\ 
Page block & 77 & 27 & 8  \\ 
Shuttle & 71 & 20 & 10  \\
Glass & 73 & 29 &  9 \\
 \hline
\end{tabular}
\end{center}
\caption{\label{fig:performance-table} A summary of the performance of the $\kappa$-detection algorithm on three real-world datasets. The threshold used in each case was chosen using Algorithm \ref{algo-threshold} except with the shuttle dataset, where $thresh = .05$ was used. In all cases, the very smallest secants were filtered out when using the SAP algorithm to calculate the $\kappa$-profile.}
\end{figure*}

Below we provide additional analysis and discussion of algorithm performance for three of the four datasets.


\subsubsection{The E. coli dataset} 
\begin{figure}
\begin{center}
\includegraphics[height=6.2cm]{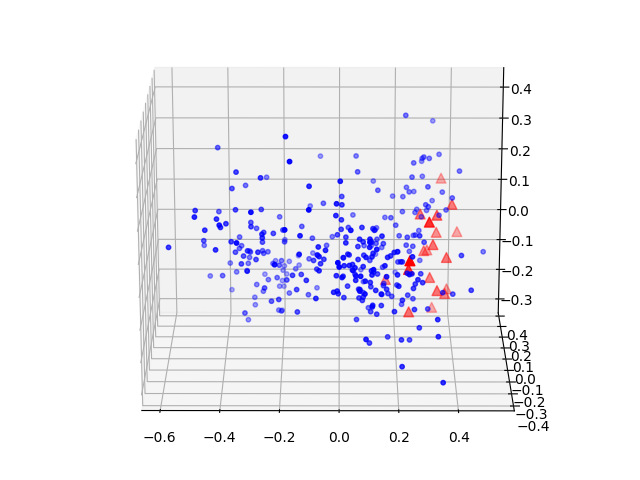}
\end{center}
\caption{\label{fig:scatter-ecoli} A projection of the points from the E. coli dataset into 3-dimensions. The points from the majority class (which contains all classes other than `om') are represented by blue circles while points from the class `om' are represented by red triangles. Note that the rare class appears to approximately sit on a 2-dimensional surface in the projection, while points from the majority class both globally and locally do not.}  
\end{figure}

\begin{figure}
\begin{center}
\includegraphics[height=6.2cm]{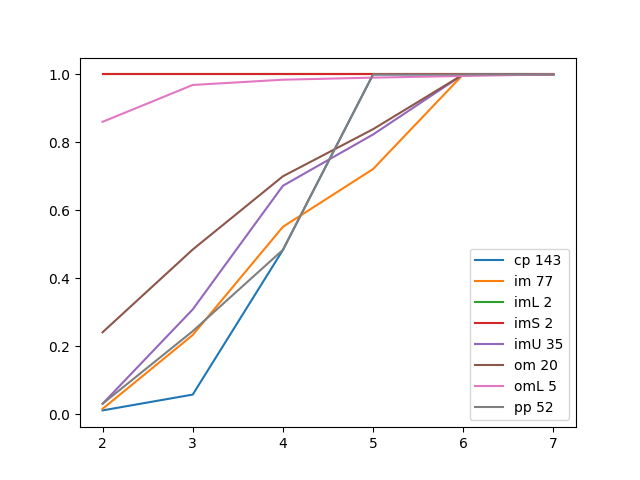}
\end{center}
\caption{\label{fig:ecoli-kappa} The $\kappa$-profiles for different classes from the E. coli dataset. The legend on the right gives the different class labels as well as the number of elements in a given class. Observe that most classes appear to be $1$ to $4$ dimesional. The rare class `om' is significantly lower dimensional than the majority class (all other classes taken as a whole). This conforms with visual inspection of the projection of this class in Figure \ref{fig:scatter-ecoli}. The three classes `imL', `imS', and `omL' have so few points that their $\kappa$-profiles are nearly trivial. We should not expect that the $\kappa$-profiles of such small samples represent the geometry of these classes.}  
\end{figure}

The E. coli dataset\footnote{https://archive.ics.uci.edu/ml/datasets/ecoli} \cite{horton1996probabilistic}, \cite{Dua:2017} consists of $8$ different classes which are related to protein localization sites on E. coli cells. The classes in this dataset vary in size, with the largest including 143 points. We choose to study the smaller class with label `om' which only contains 20 points. Figure \ref{fig:scatter-ecoli} shows a projection of the data set with the majority class (all points other than those in class `om') labeled with blue circles and all those in the rare class  `om' labeled with red triangles. From this projection of the data it appears that the rare class may very approximately sit on a 2-dimensional surface while the majority class does not. This makes this rare class a good candidate for $\kappa$-detection. The observation that the class `om' is intrinsically close to 2-dimensional is reinforced by the $\kappa$-plots for the different classes in this dataset shown in Figure \ref{fig:ecoli-kappa}. This plot gives further evidence that `om' has lower intrinsic dimension than other classes. Notice for example that the first $4$ values in the $\kappa$-profile of `om' are larger than all the classes other than those with very few points (the number of points in the class is given in the legend).

\begin{figure}
\begin{center}
\includegraphics[height=6.2cm]{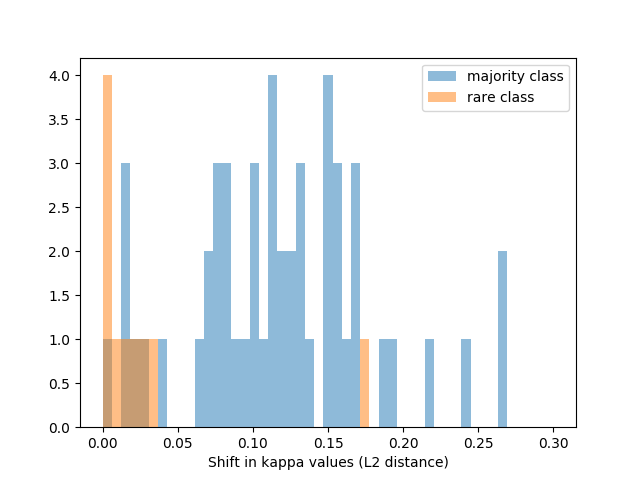}
\end{center}
\caption{\label{fig:ecoli-hist} A histogram for a random subset $Y$ of the E. coli data showing the range of values taken by $d_y$ for $y$ either in the rare class `om' (orange) or the majority class (blue). As can be seen, $d_y$ is a fairly successful mechanism for identifying the class that $y$ belongs to.}  
\end{figure}

A histogram of values for $d_y$ for one random choice of $Y$ is shown in Figure \ref{fig:ecoli-hist}. In contrast to the synthetic example, here we see that the two classes of points are not separable in the histogram. However, we do see a concentration of points from the rare class with low $d_y$ value. 

\subsubsection{The page block dataset} 
\begin{figure}
\begin{center}
\includegraphics[height=6.2cm]{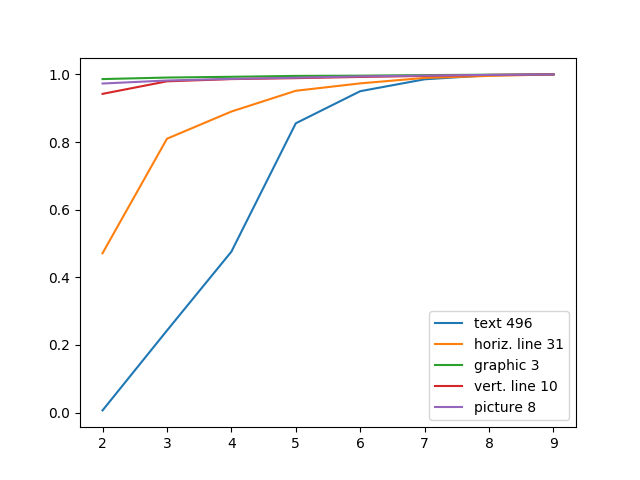}
\end{center}
\caption{\label{fig:pageblock-kappa} The $\kappa$-profiles for different classes from the page block dataset (calculated from a random sample of the entire dataset). The legend on the right gives the different class labels as well as the number of elements of a given class in the sample. As would be expected by a class with many points and a large amount of variability, the `text' class appears to be roughly $3$-dimensional. On the other hand, `graphic', `vertical line', and `picture' are close to $1$-dimensional or perhaps $2$-dimensional.}  
\end{figure}

The page block dataset\footnote{https://archive.ics.uci.edu/ml/datasets/Page+Blocks+Classification} \cite{esposito1994multistrategy,Dua:2017} consists of $5473$ data points in $\mathbb{R}^{10}$ corresponding to blocks in page layouts from $54$ distinct documents. The $10$ coordinates of the points are features related to each particular block: height, length, percentage of black pixels in the block, etc. Each point is labeled by one of $5$ different types of content contained in the block: `text', `horizontal line', `graphic', `vertical line', or `picture'. We chose the `horizontal line' class (containing $31$ points) as our rare class and took all other classes together to be the majority class. As can been seen from the $\kappa$-profiles (calculated from a subset of this dataset) in Figure \ref{fig:pageblock-kappa}, the small classes `graphic', `vertical line', and `picture' are all likely $1$-dimensional (whether this is because these classes are intrinsically $1$-dimensional or we simply don't have enough data points to estimate the dimension is unknown). The rare class `horizontal line' appears to be $1$ to $2$-dimensional while the largest class `text' is at least $3$-dimensional.

The projection of the rare class into $\mathbb{R}^3$ via PCA, Figure \ref{fig:pageblock-proj}, suggests that there are two data points that are outliers which disturb the approximate $1$-dimensionality of this dataset. After excluding these points from the set of labeled rare class points, the percentage of rare class points identified went from 70\% to 77\%, while the percentage of majority class points misidentified as rare went from 40\% to 27\%. This is an example of the effect that labeled outlier rare class points can have on the performance of the algorithm. 

\begin{figure}
\begin{center}
\includegraphics[height=6.2cm]{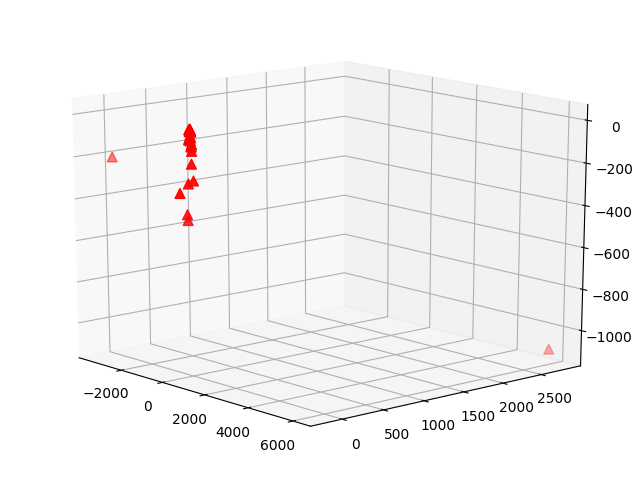}
\end{center}
\caption{\label{fig:pageblock-proj} Projection of the rare class `horizontal line' from the page block dataset into $\mathbb{R}^3$. Observe that the rare class appears to form a roughly $1$-dimensional subspace that runs vertically through the projection pictured above with two outlier points. When these outlier points were excluded, performance of the algorithm was significantly improved.}  
\end{figure}

We show a representative histogram of $d_y$ values for a single run of this dataset trained on $8$ rare-class points. We note that unlike some of the other datasets, the rare and majority classes of the page block dataset were not separable in terms of the range of values of $d_y$. Nevertheless, by discarding all unlabeled points with corresponding value $d_y$ above a well-chosen threshold, one can significantly reduce the pool of potential rare class points that must be further evaluated even when only a very small number of labeled rare class points are known.

\begin{figure}
\begin{center}
\includegraphics[height=6.2cm]{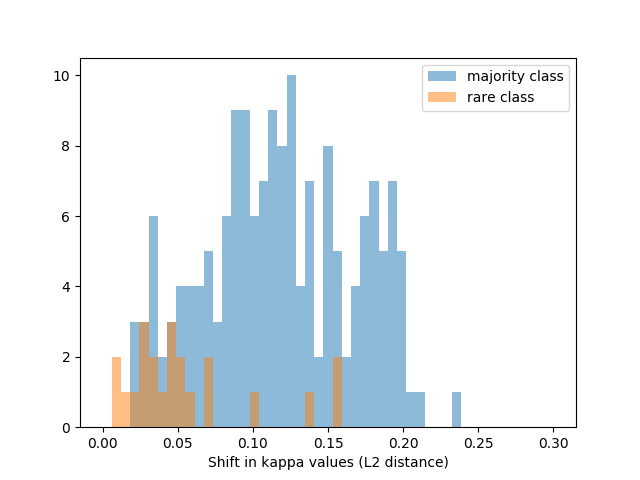}
\end{center}
\caption{\label{fig:pageblock-proj} A histogram for a random subset $Y$ of the page block dataset showing the range of values taken by $d_y$ for $y$ either in the rare class `horizontal line' (orange) or the majority class (blue). Notice that in this case, regardless of the threshold value we choose we will always misclassify some of the points. However, by choosing a threshold value close to $0.07$ for example, we can correctly identify most of the unknown points that belong to the rare class.}
\end{figure}

\subsubsection{The glass dataset}

The glass dataset\footnote{https://archive.ics.uci.edu/ml/datasets/glass+identification} \cite{Dua:2017} consists of $214$ data points in $\mathbb{R}^{10}$. The points each represent a sample of glass, the classes are given by the sample's use (window or non-window glass for example) and whether it was float processed or not. Finally the features are chemical and physical properties of the sample. We chose the class, `float processed vehicle window glass' for our rare category. This class contains 17 points out of the $214$ points.

\begin{figure}
\begin{center}
\includegraphics[height=6.2cm]{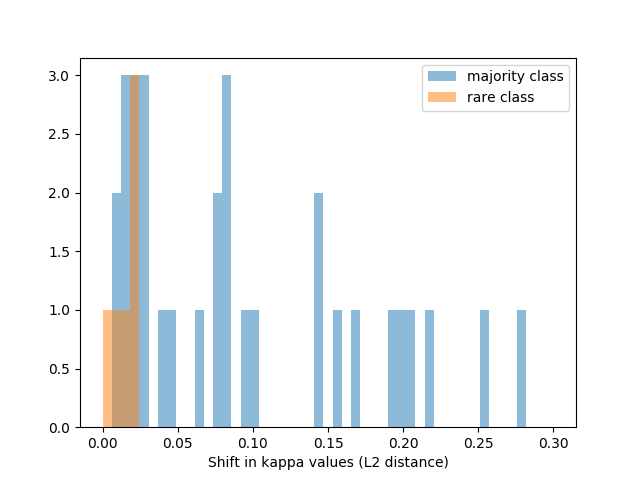}
\end{center}
\caption{\label{fig:glass-hist} A histogram of the amount by which different points from a random test set $Y$ from the glass dataset shifted the $\kappa$-profile. For this dataset, a significant fraction of the majority class are situated such that inclusion into the rare class does not disturb its geometry.}  
\end{figure}

\begin{figure}
\begin{center}
\includegraphics[height=6.2cm]{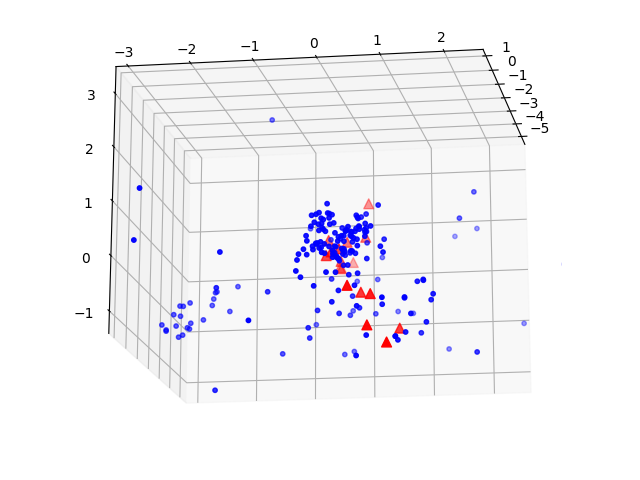}
\end{center}
\caption{\label{fig:glass-proj} Projection of the glass dataset into $\mathbb{R}^3$. The rare class points `float processed vehicle window glass' are shown as red triangles while the majority class points are blue circles. Notice that, at least in this projection, the rare and majority class are not separable.}  
\end{figure}

Of all the datasets that we tested our algorithm on, this dataset had the highest rate of misclassification of majority class points as rare class points. A representative histogram of $d_y$ values for one run of $\kappa$-detection on this dataset is shown in Figure \ref{fig:glass-hist}. Here we see that a significant number of the unlabeled majority are situated such that their inclusion into the rare class minimally disturbs its $\kappa$-profile. This is supported by inspection of this dataset projected into $\mathbb{R}^3$ using PCA (Figure \ref{fig:glass-proj}). At least in this projection, it appears that a cluster of points from the majority class sit along the same approximate 2-dimensional surface on which the rare class sits. This is another reminder that the $\kappa$-detection algorithm is limited by the geometry of the points it is given. If the data manifold for two classes coincide in a significant way, we should expect limited classification accuracy for points in those regions.

\section{Conclusion}
\label{sec:conclusion}

In this paper we propose a new approach to the rare-category detection problem, which, given a small set of labeled points from a rare category, finds others based on geometric/dimensionality considerations.

There are a number of directions which would be interesting to explore in the future.
\begin{itemize}
    \item Visual inspection indicates that many of the errors made by the $\kappa$-detection algorithm are due to noise. At the moment the only tool we have applied to address this is to discard small secants. It would be useful to develop more sophisticated methods.
    \item Are there more appropriate norms for measuring change in the $\kappa$-profile? Are there certain coordinates in the $\kappa$-profile that we should pay particular attention to?
    \item How well does the $\kappa$-detection algorithm contribute to ensemble techniques in the case when majority and rare classes have the same underlying dimension?
    \item While in this paper we focus on using $\kappa$-detection to make a binary classification of an unlabeled point as belonging to a rare class or not, one could also use the value $d_y$ directly. In a future work, we plan to investigate how the values $d_y$ can be used to calculate probabilities that the point $y$ belongs to a given data manifold.
\end{itemize}

\bibliography{refs}
\bibliographystyle{IEEEtran}


\end{document}